\documentclass[conference]{IEEEtran}
\IEEEoverridecommandlockouts
\usepackage{cite}
\usepackage{amsmath,amssymb,amsfonts}
\usepackage{algorithmic}
\usepackage{graphicx}
\usepackage{textcomp}
\usepackage{mathrsfs}
\usepackage{xcolor}
\usepackage{caption}
\usepackage{subcaption}

\def\BibTeX{{\rm B\kern-.05em{\sc i\kern-.025em b}\kern-.08em
    T\kern-.1667em\lower.7ex\hbox{E}\kern-.125emX}}
\def\endthebibliography{%
  \def\@noitemerr{\@latex@warning{Empty `thebibliography' environment}}%
  \endlist
}
\begin{document}
\title{Deep Reinforcement Learning for Community Battery Scheduling under Uncertainties of Load, PV Generation, and Energy Prices
\thanks{This work was supported in part by the Australian Research Council (ARC) Discovery Early Career Researcher Award (DECRA) under Grant DE230100046.}
}
\author{\IEEEauthorblockN{Jiarong Fan\textsuperscript{1}, Hao Wang\textsuperscript{1,2*}}
\IEEEauthorblockA{
    \textsuperscript{1}Department of Data Science and AI, Faculty of IT, Monash University, Melbourne, VIC 3800, Australia \\
    \textsuperscript{2}Monash Energy Institute, Monash University, Melbourne, VIC 3800, Australia}
\thanks{*Corresponding author: Hao Wang (hao.wang2@monash.edu).}
}

\maketitle

\begin{abstract}
In response to the growing uptake of distributed energy resources (DERs), community batteries have emerged as a promising solution to support renewable energy integration, reduce peak load, and enhance grid reliability. This paper presents a deep reinforcement learning (RL) strategy, centered around the soft actor-critic (SAC) algorithm, to schedule a community battery system in the presence of uncertainties, such as solar photovoltaic (PV) generation, local demand, and real-time energy prices. We position the community battery to play a versatile role, in integrating local PV energy, reducing peak load, and exploiting energy price fluctuations for arbitrage, thereby minimizing the system cost. To improve exploration and convergence during RL training, we utilize the noisy network technique. This paper conducts a comparative study of different RL algorithms, including proximal policy optimization (PPO) and deep deterministic policy gradient (DDPG) algorithms, to evaluate their effectiveness in the community battery scheduling problem. The results demonstrate the potential of RL in addressing community battery scheduling challenges and show that the SAC algorithm achieves the best performance compared to RL and optimization benchmarks.
\end{abstract}

\begin{IEEEkeywords}
Community battery, solar photovoltaic, energy management, uncertainties, reinforcement learning.
\end{IEEEkeywords}

\section{Introduction}
In recent years, community batteries have been increasingly recognized as a pivotal solution to the challenges arising from the growing uptake of distributed energy resources (DERs), in particular residential solar photovoltaic (PV) systems. Defined as a system with community ownership, community batteries generate collective socio-economic benefits, such as hosting more renewable energy, increasing higher self-consumption of renewables, and reducing electricity bills \cite{koirala2018community}.
Different from behind-the-meter batteries, community batteries can provide shared storage for multiple households. Such a battery sharing paradigm has been demonstrated to be more cost-effective than residential behind-the-meter batteries in existing studies, such as \cite{zhao2019virtual, zhao2017pricing}. In Australia, both federal and state governments, such as Victoria and Western Australia, have been supporting the development of community batteries, which are expected to store and share excess solar power in residential communities and shave peak load on distribution networks.

Community batteries can serve multiple roles in the distribution system by integrating local renewables, optimizing energy consumption, and enhancing the reliability of distribution networks. The key to unlocking the value of community batteries lies in the scheduling of charge and discharge of community batteries under uncertainties in the systems. Therefore, developing effective scheduling strategies for community batteries becomes a critical area of research. 

A large body of literature has explored various optimization techniques for community battery scheduling. For example, a model predictive control method was applied in \cite{abraham2023mpc} to optimize an energy system of two townhouses with grid-connected PV systems and a community battery. A receding predictive horizon method was proposed in \cite{wolfs2012receding} for periodic optimization of community battery systems. Specifically, this method involves the load and PV generation prediction, followed by updating the optimal scheduling of community battery systems hourly to adapt to changing future predictions. Similar methods for community battery system optimization was presented in another study \cite{pezeshki2014model}. The optimal operation of an energy community was studied in 
\cite{stentati2022optimization}, which formulated an optimization problem to compute the battery charging/discharging policies and the set points of flexible loads and controllable generators. In addition to minimizing costs or optimizing economic performance, some other studies considered different objectives. For example, \cite{van2020low} used a community battery to reduce grid congestion and control the voltage and current within the network. In \cite{wang2017pv}, the battery control problem was formulated as a linear optimization problem and solved by a receding horizon method. In addition, different energy storage options and optimization objectives (e.g., energy minimization, cost minimization and electricity self-sufficiency) were considered. Another study using multi-objective optimization method was presented in \cite{wang2023community} to minimize voltage deviation for distribution network operators, while maximizing the utilization of DERs on the demand side.

The existing studies above \cite{abraham2023mpc,wolfs2012receding,pezeshki2014model,stentati2022optimization,van2020low,wang2017pv,wang2023community} primarily focused on optimization techniques, which can contribute to the operation of community batteries in systems with load and PV generation. However, these methods often fall short in the face of various uncertainties in the operation of community battery systems. Specifically, renewable energy generation, such as residential PV generation, fluctuates, electricity demand varies over time, and network conditions are difficult to predict. These uncertainties can impede the performance of optimization-based methods, as they usually rely on the quality of forecasts to a large extent. More effective methodologies are needed to deal with uncertainties when optimizing community battery scheduling.

Handling the aforementioned uncertainties and making community batteries more adaptive to time-varying conditions is where reinforcement learning (RL) emerges as a promising solution. RL has gained significant traction across a wide range of applications. For example, RL has used in optimizing energy storage bidding, electric vehicle (EV) charging, and energy system operations \cite{wang2018energy,badoual2021learning,li2023deep,sultanuddin2023development,cao2021smart,alfaverh2023optimal,kosuru2021reinforcement,rostmnezhad2023power,liu2023learning,lix2023deep}. Studies have demonstrated RL's effectiveness in developing energy storage bidding strategies \cite{wang2018energy,badoual2021learning,li2023deep} in electricity markets using different RL algorithms, such as Q learning, supervised actor-critic, and twin-delayed deep deterministic policy gradient (TD3) algorithms. The above studies showed that RL can effectively handle uncertainties in the market environment. RL algorithms have also been applied to develop EV charging and discharging coordination strategies, including smart charging policy using double Q-learning \cite{sultanuddin2023development} and customized actor-critic learning \cite{cao2021smart}, vehicle-to-grid control using deep deterministic policy gradient (DDPG) algorithm \cite{alfaverh2023optimal}, vehicle-to-vehicle energy exchange using multi-agent DDPG algorithm \cite{fan2023marl}. These studies have validated the effectiveness of RL in optimizing EV coordination against uncertainties in EV charging behaviors. The operation of energy storage systems in residential communities or microgrids has been studied using RL algorithms, such as Q-learning for a PV and battery connected microgrid \cite{kosuru2021reinforcement}, Q-learning for both thermal and battery energy storage systems in campus buildings \cite{rostmnezhad2023power}, transformer-based DDPG for virtual storage rental service \cite{lix2023deep}, and multi-agent DDPG algorithm for a shared energy storage system in a residential community \cite{li2023deep}.
Despite RL's potential, it relies on extensive exploration of the environment. Inadequate exploration can result in local optimum. The above studies did not sufficiently enhance the exploration ability of the RL algorithms, potentially leading to low data efficiency and performance.

In this paper, we present a deep reinforcement learning (DRL) strategy for scheduling a community battery system in the face of various uncertainties, such as PV generation, local load, and market prices. We leverage the community battery's versatile capability to integrate local PV energy, shave peak loads, and exploit price fluctuations for energy arbitrage. By utilizing RL and enhancing its exploration ability, we develop a soft actor critic (SAC) algorithm to improve the training and performance of RL in scheduling community batteries under uncertainties. Furthermore, there are several prominent algorithms, including both model-based optimization and RL algorithms, renowned for their efficacy in various energy system applications. We also aim to implement these algorithms and compare their effectiveness with our SAC algorithm in the context of community battery scheduling problem.

The contributions of this paper are as follows.
\begin{itemize}
    \item \textit{Community battery scheduling under uncertainties:} We use SAC algorithm to enhance community battery scheduling amidst uncertainties in solar PV generation, load, and energy prices, while satisfying system constraints.
    \item \textit{Noisy network for better exploration in RL training:} We employ a noisy network as opposed to action noise, which can foster RL exploration to accelerate convergence. Numerical results show that the proposed approach improves the convergence during RL training.
    \item \textit{Comparative study of different algorithms:} We conduct comparative studies of model-based optimization and model-free RL algorithms, including DDPG, proximal policy optimization (PPO), and SAC algorithms, each representing different RL paradigms, such as on/off-policy and deterministic/stochastic policy. Numerical results show the superior performance of the proposed SAC algorithm in scheduling community battery. 
\end{itemize}

The remainder of this paper is organized as follows. Section \ref{sec:problem} formulates the community battery scheduling problem. Section \ref{sec:mdp} presents the Markov decision process formulation for the studied problem. Section \ref{sec:rl} presents the SAC algorithm. Section \ref{sec:eval} discusses the numerical results and findings. Section \ref{sec:con} concludes this paper.

\section{Problem Formulation}\label{sec:problem}
\begin{figure}[t]
    \centering
    \includegraphics[width=0.9\linewidth]{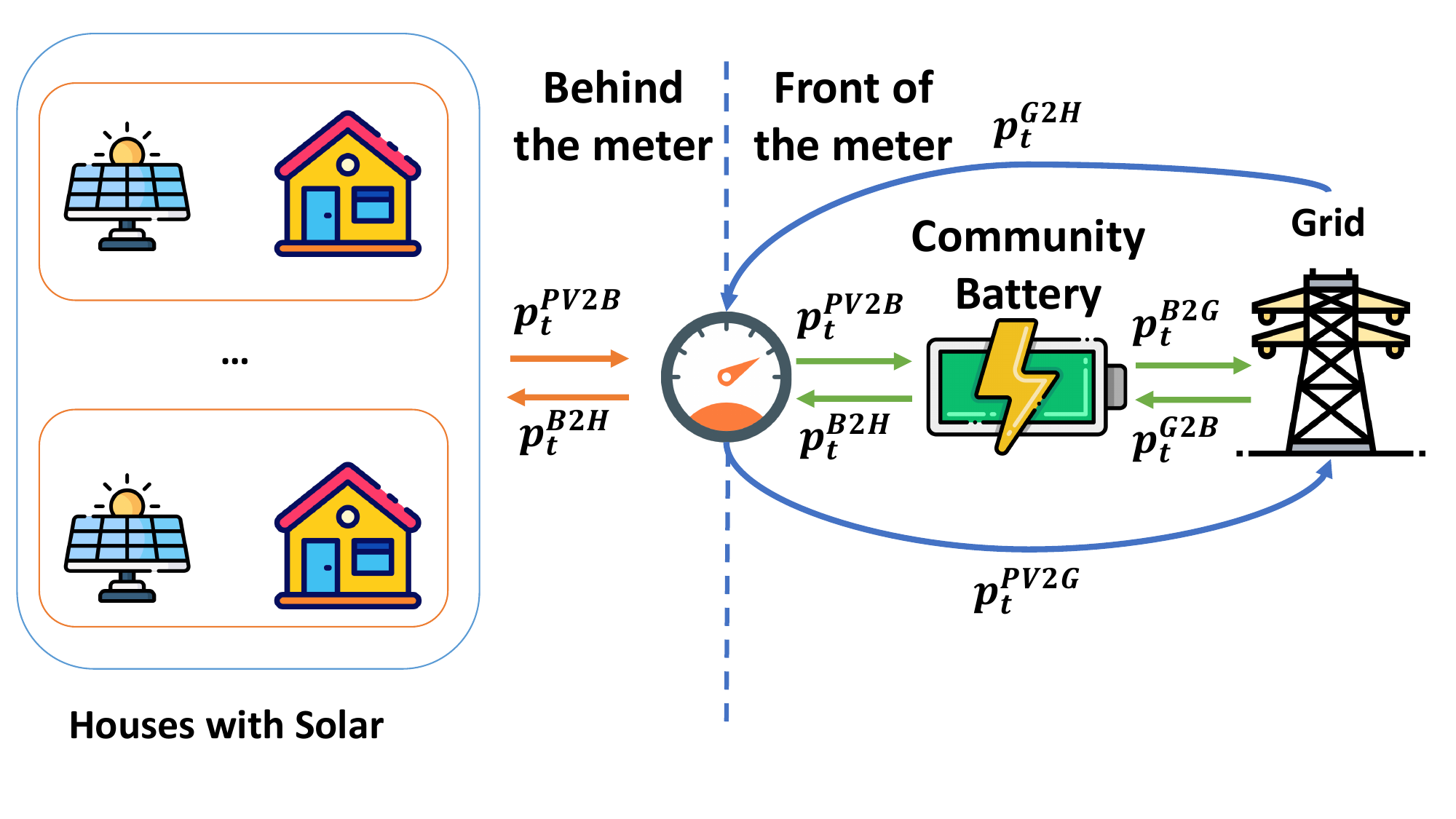}
    \caption{Overview of the community battery system with residential homes and solar PV in a grid-connected scenario.}
    \label{system}
\end{figure}

Fig.~\ref{system} illustrates a schematic overview of the community battery system under study. Residential homes with PV panels are connected behind the meter, and they can buy and sell energy with both the community battery and the grid. The community battery is positioned in front of the meter, enabling its interactions with multiple homes. The community battery aims to minimize the system energy costs by scheduling its charging and discharging within an operational horizon $\mathcal{T}$ with $\Delta t$ as the time interval. For example, the community battery stores surplus solar energy and charges grid power when the price is relatively low. Conversely, it discharges to supply energy to homes and also export energy back to the grid for additional benefits. In the following, we will present the system model in detail.

\subsubsection{Community Battery Model}
The community battery is a main component in this system, supporting bidirectional power flow. We denote $p_t^\text{ch}$ and $p_t^\text{disch}$ as the charge and discharge power of the community battery at $t \in \mathcal{T}$. To avoid charge and discharge at the same time, we introduce a binary variable $z_{t}$. The constraints for charge and discharge are as follows
\begin{align}
    &0 \leq p_{t}^\text{ch} \leq (1-z_{t})\bar{P}^\text{ch}\label{xt_con},\\
    &0 \leq p_{t}^\text{disch} \leq z_{t}\bar{P}^\text{disch} \label{xdt_con},\\
    & z_{t} \in \{0,1\} \label{z_it},
\end{align}
where $\bar{P}^\text{ch}$ and $\bar{P}^\text{disch}$ are the maximum charging and discharging power of the community battery. The binary variable $z_{t}$ can choose $0$ and $1$ to enable charge and discharge so as to avoid charge and discharge happen simultaneously.

With the community battery charge and discharge, the battery energy level $E_t$ evolves accordingly. The energy level dynamics can be modeled as
\begin{align}
        &E_{t} = E_{t-1}+ p_{t}^\text{ch} \eta^{\text{ch}} \Delta t -  \frac{p_{t}^\text{disch}\Delta t} {\eta^{\text{disch}}} \label{eit},\\
        &E^\text{min}\leq E_{t} \leq E^\text{max}, \label{eit_con}
\end{align}
where $\eta^\text{ch}$ and $\eta^\text{disch}$ are the charging/discharging efficiency parameters of the community battery. In \eqref{eit_con}, $E^\text{min}$ and $E^\text{max}$ are the minimum/maximum energy levels allowed for the community battery.

Charging and discharging batteries often incur battery degradation, which can be modeled by the degradation cost. In our work, we adopt the energy throughput equivalent method \cite{wu2022optimal} to assess the battery degradation cost by determining the equivalent  full cycle $EFC_t$ at time $t$, which counts the energy throughput as a proportion of battery capacity $E^\text{cap}$. According to the energy throughput equivalent method, the battery degradation $AGE_t^\text{cyc}$ at time $t$ is shown as follows
\begin{align}
    &EFC_{t} = 0.5 \cdot \frac{\lvert p_{t}^\text{ch} \eta^{\text{ch}} \Delta t -  \frac{p_{t}^\text{disch}\Delta t} {\eta^{\text{disch}}} \rvert}{E^\text{cap}}, \label{EFC}\\
    &AGE_{t}^\text{cyc} = \frac{EFC_{t}}{L^\text{cyc}}, \label{AGE}
\end{align}
in which $L^\text{cyc}$ denotes the community battery life cycle. The value 0.5 corresponds to the average half-cycle within the battery's lifespan.

\subsubsection{Energy Balance}
In this system, we consider one residential homes with their PV panels as an entity. The total solar PV power generation $P_t^\text{PVgen}$ will be first supplied to meet the total residential load $P_t^\text{home}$ for self-consumption. The net load of homes is calculated as $P_t^\text{net} = P_t^\text{home} - P_t^\text{PVgen}$. A positive value of $P_t^\text{net}$ signifies insufficient PV power generation, necessitating extra power supply from the battery $p_t^\text{B2H}$ and grid $p_t^\text{G2H}$. Conversely, a negative $P_t^\text{net}$ indicates that there is surplus PV power. This excess PV power can then be directed to the community battery $p_t^\text{PV2B}$ and sold back to the grid $p_t^\text{PV2G}$. The energy balance constraints can be expressed as
\begin{align}
    &  p_t^\text{G2H} + p_t^\text{B2H} = (1-u_t)P_t^\text{net},\label{net_power}\\ 
    &  -(p_t^\text{PV2G} + p_t^\text{PV2B}) \geq u_t P_t^\text{net},\\
    & p_t^\text{ch} = p_t^\text{PV2B} + p_t^\text{G2B},\label{pchb}\\
    & p_t^\text{disch} = p_t^\text{B2H} + p_t^\text{B2G}, \label{pdischb}\\
    & u_t \in \{0,1\},\\
    & p_t^\text{G2H}, p_t^\text{B2H}, p_t^\text{PV2G}, p_t^\text{PV2B},  p_t^\text{B2G}, p_t^\text{G2B} \geq 0, \label{allgeq}
\end{align}
in which the net load must be met by the battery $p_t^\text{G2H}$ and the grid $p_t^\text{B2H}$. The excessive PV power can be supplied to the battery $p_t^\text{PV2B}$ and the grid $p_t^\text{PV2G}$ with curtailment allowed. We specify $p_t^\text{G2B}$ and $p_t^\text{B2G}$ as the battery charging power from grid and discharging power to grid, respectively. The binary variable $u_t$ is used to avoid selling power from the PV and buying power for homes occurring simultaneously.

Moreover, the entire system, including residential homes, PV energy, and the community battery, should satisfy the grid capacity constraints, which are shown as
\begin{align}
    & 0 \leq p_t^\text{G2H} + p_t^\text{G2B} \leq G^\text{max},\label{g2}\\
    & 0 \leq p_t^\text{B2G} + p_t^\text{PV2G} \leq G^\text{max}, \label{2g}
\end{align}
where $G^\text{max}$ is the maximum power import and export capacity between the home and the grid. Note that $G^\text{max}$ is usually greater than the maximum home load. The import and export capacities can be different.

\subsubsection{Objective and Optimization Formulation}
The objective of the problem is to minimize the total cost of the system, which includes energy cost and the community battery degradation cost. Specifically, the optimization problem is presented as
\begin{align}
\begin{split}
&\min ~~ \!\!\!\!\sum_{t\in \mathcal{T}} \Bigl( (p_t^\text{G2H} + p_t^\text{G2B})\kappa_t^\text{buy}\Delta t \\
&+ (p_t^\text{B2G} + p_t^\text{PV2G})\kappa_t^\text{sell}\Delta t 
+ AGE_{t}^\text{cyc} \kappa^\text{batt} \Bigr) \\
&\text{s.t.}~~\eqref{xt_con} - \eqref{eit_con}~\text{and}~\eqref{net_power} - \eqref{2g},
\end{split}
\label{obj}
\end{align}
where $\kappa^{\text{buy}}_t$, $\kappa^{\text{sell}}_t$ and $\kappa^{\text{batt}}_t$ are energy selling price, energy purchasing price and battery degradation cost coefficient.

\section{Markov Decision Process}\label{sec:mdp}
This section formulates a Markov decision process (MDP) for the community battery scheduling problem, as described in Section~\ref{sec:problem}. The MDP formulation mainly consists of state, action, state transition, and reward. We consider the operator of the community battery as an RL agent, and the agent utilizes state information to take actions to maximize the reward.

\textbf{State:}
The state describes the current situation of the system and critical information for computing rewards. For our work, the state includes information regarding the battery's status as well as exogenous energy data. The states in the MDP generally fall into two categories: those that are contingent on the agent's actions, namely action-related states $s_{t}^a$, and those that are independent of the agent's actions, which are independent states $s_{t}^\text{ind}$. Action-related states, which are directly affected by agents' actions, include the energy level of the community battery $E_{t}$. Conversely, independent states describe exogenous information, such as the energy purchasing price $\kappa_t^\text{buy}$, energy selling price $\kappa_t^\text{sell}$, residential load $P_t^\text{home}$, and solar PV generation $P_t^\text{PVgen}$. The state observed by the agent at time $t$ can be modeled as
\begin{align}
    &s_{t}^a {=} \{E_{t}\} \label{state_a},\\
    &\!s_{t}^\text{ind} {=} \{\kappa_{t}^\text{buy},\kappa_t^\text{sell}, P_t^\text{PVgen}, P_t^\text{home}\}, \label{state_ind}\\
    &s_{t} {=} \{s_{t}^a, s_{t}^\text{ind}\}.  \label{state1}
\end{align}

\textbf{Action:} 
The action of the agent is the decision of the community battery operator, including the charging and discharging power and their interactions with PV, residential load, and the grid. The action can be expressed as
\begin{align}
    a_{t} = \{p_{t}^\text{ch},p_t^\text{disch}, a_{t}^\text{grid}\},
    \label{action}
\end{align}
in which the battery charging power $p_t^\text{ch}$ is sourced from both the grid and solar energy. The discharging power $p_t^\text{disch}$ is directed to the grid and homes. We introduce action $a_t^\text{grid}$ to represent the fraction of energy, either charged from or discharged to the grid. Once the charging/discharging power and fraction are determined, all variables in \eqref{pchb} and \eqref{pdischb} can be calculated.

\textbf{State Transition:} 
During the battery charging/discharging, the action-related state $s_t^a$ follows the battery dynamics \eqref{eit}. The independent states can be updated using real-time data.

\textbf{Reward:}
The goal of the RL agent is to optimize the cumulative reward. Consequently, the design of the reward is intrinsically linked to the specific objectives \eqref{obj}. In our work, the goal is to leverage the community battery to facilitate load shifting and minimize the overall system cost, which comprises both the energy cost and the battery degradation cost. Within the framework of RL, these costs are treated inversely, with negative costs translated to rewards. The reward associated with system costs, denoted as $R_t^\text{cost}$, can be expressed as
\begin{align}
        &R_t^\text{ec} = (p_t^\text{G2H} + p_t^\text{G2B})\kappa_t^\text{buy}\Delta t + (p_t^\text{B2G} + p_t^\text{PV2G})\kappa_t^\text{sell}\Delta t, \label{ec}\\
        &R_t^\text{batt} = AGE_{t}^\text{cyc} \kappa^\text{batt}, \label{batt_r} \\
        &R_t^\text{cost} = -(R_t^\text{ec} + R_{t}^\text{batt}), \label{cost_r}
\end{align}
where $R_t^\text{ec}$ and $R_t^\text{batt}$ are energy cost and battery degradation cost. The studied system not only minimizes the total cost but also aims to satisfy the network capacity constraint. Therefore, we set a penalty $R_t^\text{penalty}$ for capacity constraint violation. The final reward can be expressed as
\begin{align}
        &R_t = \xi R_t^\text{cost} + (1-\xi)R_t^\text{penalty}, \label{eq_finalreward}
\end{align}
where $\xi$ specifies the trade-off between the total cost and the capacity constraint.

\section{Proposed RL Method}\label{sec:rl}

\begin{figure}[t]
    \centering
    \includegraphics[width=0.9\linewidth]{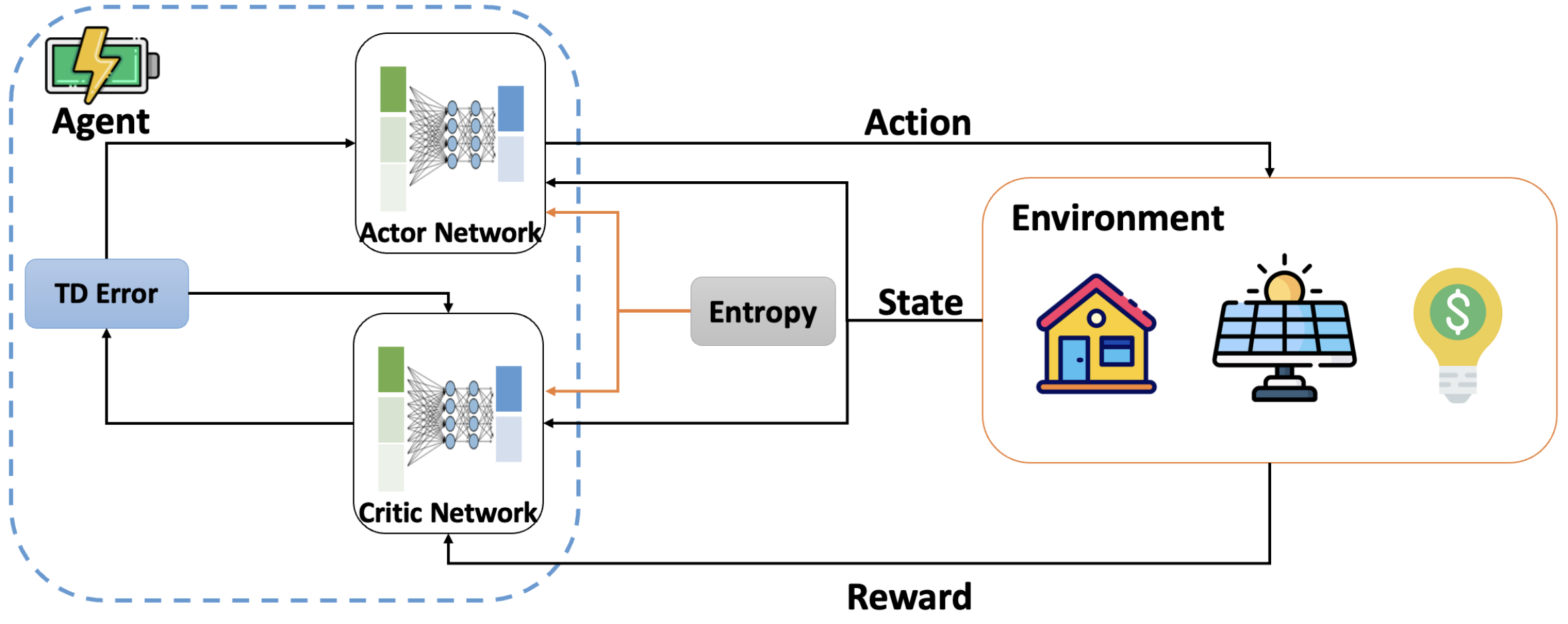}
    \caption{The framework for soft actor critic algorithm.}
    \label{sac}
\end{figure}

In this section, we present the RL algorithm, which learns a community battery scheduling strategy to optimize the cumulative reward in Eq. \eqref{eq_finalreward}. In traditional Q-learning, the objective is to find an optimal deterministic policy. In contrast, SAC, as a promising RL algorithm, deviates from the traditional Q-learning by aiming to learn a stochastic policy. We employ SAC because stochastic policies usually provide better exploration of the state space, especially in complex and high-dimensional environments. The framework of SAC is illustrated in Fig \ref{sac}, which is based on the actor critic (A2C) framework. In the A2C framework, the actor network approximates the state-value function and generates the action of the community battery operator, while the critic approximates the action-value function, evaluating the actor's actions, and minimizes the TD-error via the critic network.

SAC's objective is rooted in the maximum entropy RL framework. In this framework, the objective is not only to maximize the expected return but also to maximize the entropy of the policy. This effectively encourages more explorative policies. In the maximum-entropy RL framework, the regularized state-value function $V_\text{soft}(\textbf{s})$ is defined as
\begin{align}
        \begin{split}
        & V_\text{soft}(\textbf{s}) = \mathbb{E}_{\tau \sim \pi}\big[
        \sum_{t=0}^T\gamma(R_{t} + \alpha H(\pi(\cdot|\textbf{s}_t)))|\textbf{s}_0=s
        \big], \label{state_v}
        \end{split}
\end{align}
where $H(\pi(\cdot|\textbf{s}_t)))$ is the entropy for a probabilistic policy $\pi$ at state $\textbf{s}_t$, $\pi$ is the mapping between state and action, and $\gamma$ is discount factor. The learning rate of the value function is $\alpha$. The action-value function $Q_\text{soft}(\textbf{s}, \textbf{a})$ is defined as
\begin{align}
  \begin{split}
  & Q_\text{soft}(\textbf{s}, \textbf{a}) = \mathbb{E}_{\tau \sim \pi}
  \big[ \sum_{t=0}^T \gamma R_t\\ &+ \alpha \sum_{t=0}^T \gamma H(\pi(\cdot|\textbf{s}_t)))|\textbf{s}_0=s, \textbf{a}_0=a
  \big].
  \end{split}
\end{align}
The relationship of above two value functions can be represented as
\begin{align}
    & V_\text{soft}(\textbf{s}_t) = \mathbb{E}_{a_t \sim \pi}[Q_\text{soft}(\textbf{s}_t,\textbf{a}_t)] + \alpha H(\pi(\cdot|\textbf{s}_t))).
\end{align}

Note that when the action-value function $Q_\text{soft}$ converges to $Q_\text{soft}^*$, the optimal policy $\pi^*$ also achieves the optimal value $V_\text{soft}^*$. The value functions is updated according to \cite{haarnoja2018soft}. 

Additionally, to enhance exploration, we introduce parameter noises to the neural network in the SAC algorithm. We refine the original algorithm by integrating a noisy network, as suggested in \cite{plappert2017parameter}. This modification entails replacing the conventional linear layer with a noisy linear layer, which can be characterized as
\begin{align}
    & Y = (\nu^\theta + \sigma^\theta \odot \epsilon^\theta)X + (\nu^b + \sigma^b \odot \epsilon^b),
\end{align}
where $\epsilon = [\epsilon^\theta, \epsilon^b]$ denotes matrices of noise randomly sampled, each exhibiting a zero mean and fixed statistics. Furthermore, $\nu = [\nu^\theta, \nu^b]$ and $\sigma = [\sigma^\theta, \sigma^b]$ represent the neural network's learnable parameters for weights and biases, respectively. Within the noisy network framework, the agent avoids the typical $\epsilon$-greedy policy; instead, it acts greedily based on a network that employs noisy linear layers for decision-making.

\section{PERFORMANCE EVALUATION} \label{sec:eval}

\subsection{Simulation Setup and Benchmarks}
In our numerical experiments, we consider a community model comprising $60$ residential homes, each equipped with a 2 kWp solar PV system. These residential homes together with their solar energy are considered as a group of local load and generation interacting with both the community battery and grid. The community battery is located in front of the meter with a capacity of 500 kWh and a charge/discharge rate of 250 kW. We assume that the community battery purchases energy based on the retail Time-of-Use (TOU) price and sells surplus energy back to the grid at wholesale prices. The real-time wholesale prices can be obtained from the National Electricity Market (NEM) in Australia, as provided by the Australian Energy Market Operator (AEMO) \cite{aemo_dashboard}.

For scheduling the community battery, we utilize the SAC with a stochastic policy, as it can introduce greater randomness in the decision-making process. Comparative analysis includes model-based optimization, DDPG and PPO algorithms. The model-based optimization is based on receding predictive horizon approach \cite{wolfs2012receding}, which leverages the future data prediction to assist decision making. Moreover, DDPG and PPO are representative deterministic policy and on-policy RL algorithms, respectively.

\subsection{Simulation Results and Discussion}
\begin{figure}[t]\centering
     \begin{subfigure}[b]{0.23\textwidth}
         \includegraphics[width=4.5cm]{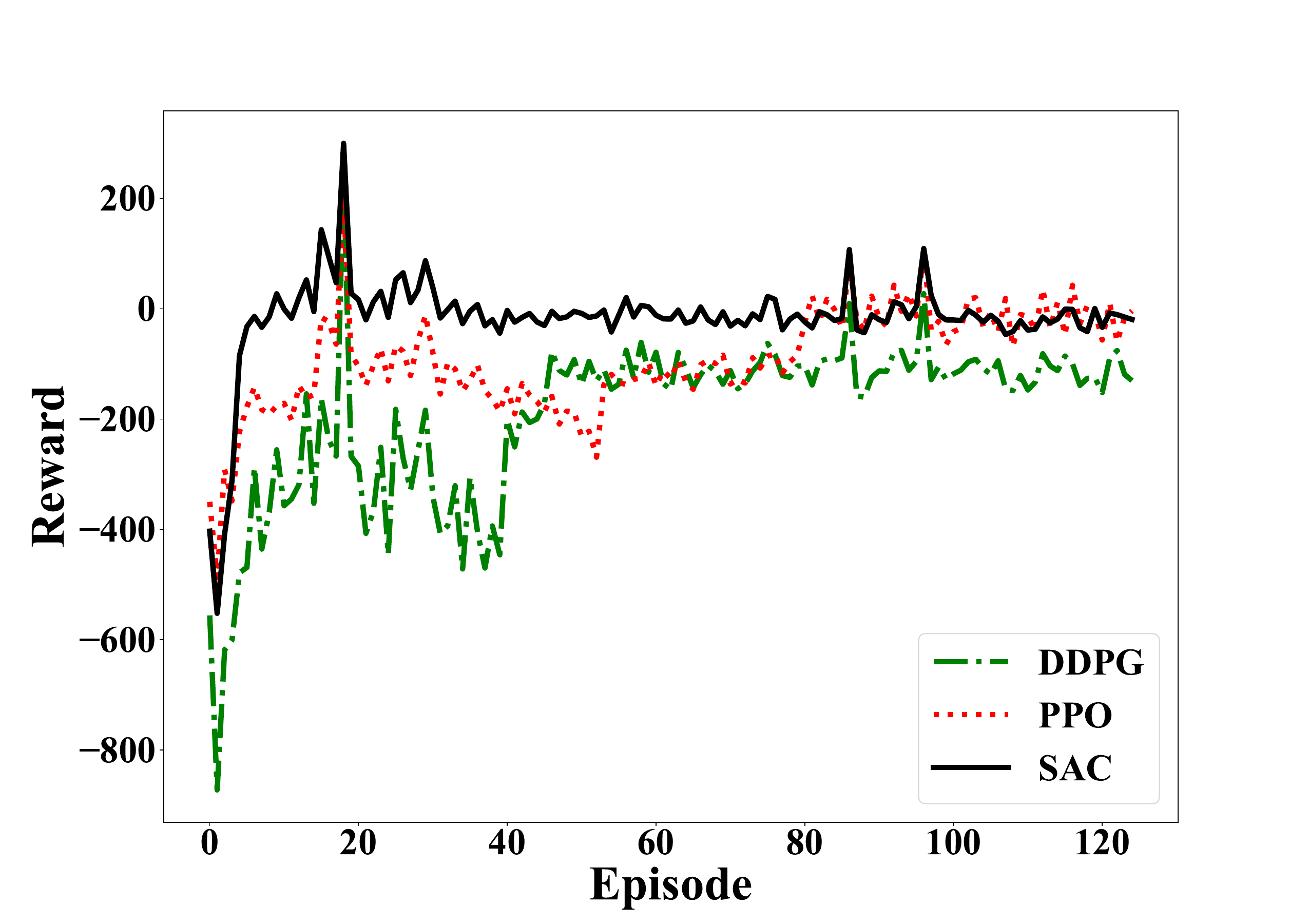}
         \caption{DDPG, PPO and SAC}
         \label{reward}
     \end{subfigure}
     \begin{subfigure}[b]{0.23\textwidth}
         \includegraphics[width=4.5cm]{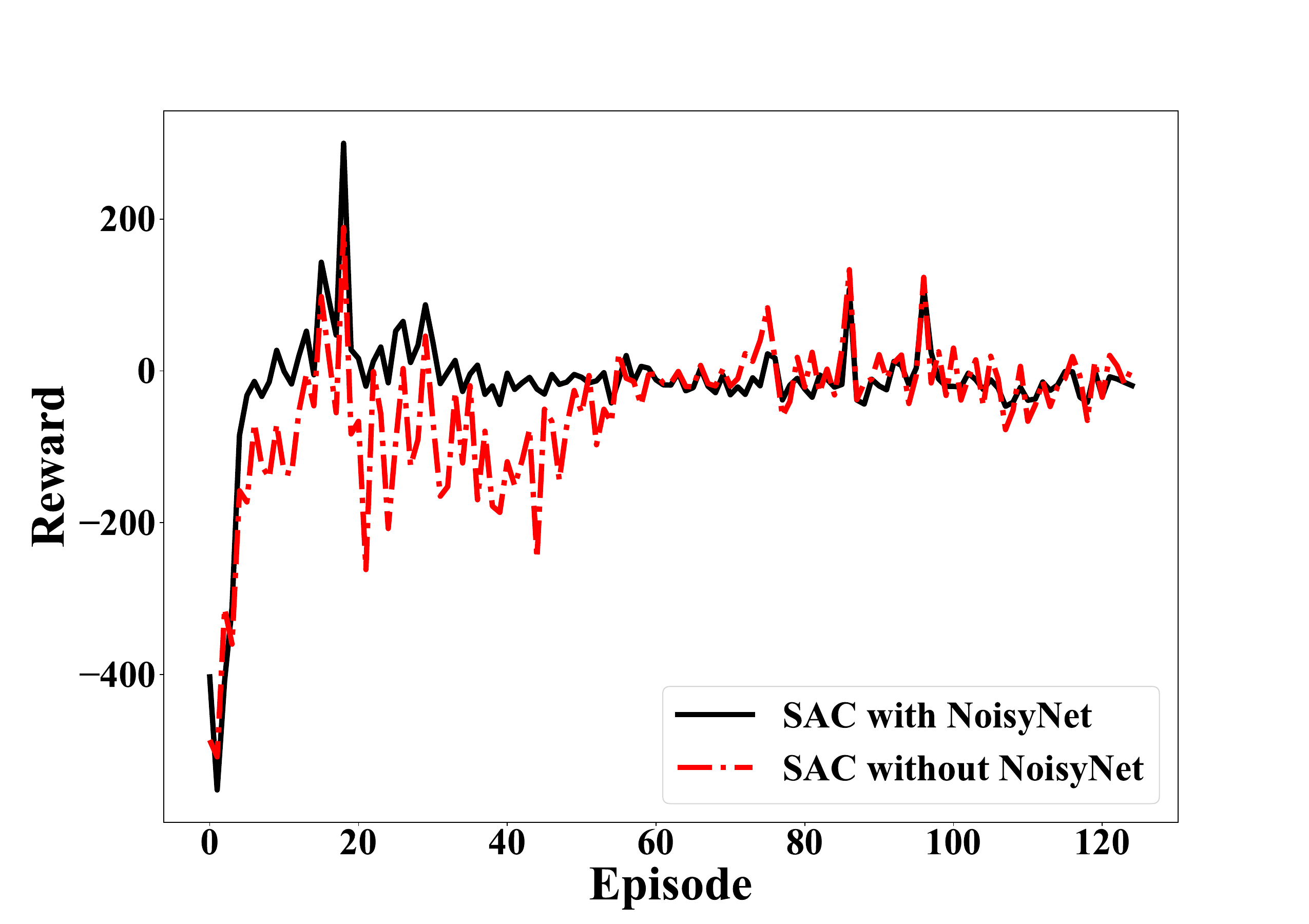}
         \caption{NoisyNet}
         \label{reward_noise}
     \end{subfigure}
        \caption{The reward convergence comparison across different algorithms.}
    \label{baseline}
\end{figure}

Fig. \ref{reward} shows the convergence of rewards for three RL algorithms, all of which converge after 90 episodes. Both DDPG and SAC algorithms exhibit faster convergence rates. However, the reward associated with DDPG is lower, likely due to its deterministic policy's propensity for local optima, caused by insufficient exploration. In contrast, PPO achieves a competitive reward compared to SAC, despite PPO's slower convergence rate, likely caused by the inherently lower data efficiency of on-policy RL methods. Nonetheless, the integration of a stochastic policy with a maximum entropy framework proves advantageous, enhancing both the convergence and the cumulative reward.
Our experiment further assesses the effectiveness of the NoisyNet. Figure \ref{reward_noise} depicts the convergence of rewards for the SAC algorithm with and without NoisyNet. The reward curves reveal that SAC integrated with NoisyNet achieves a higher convergence rate, compared to SAC without NoisyNet. By introducing parameter noises within the neural network, NoisyNet broadens the diversity of the generated samples, thereby accelerating the learning process leading to more effective policies.

\begin{figure*}[t]
    \centering
    \includegraphics[width=0.95\linewidth]{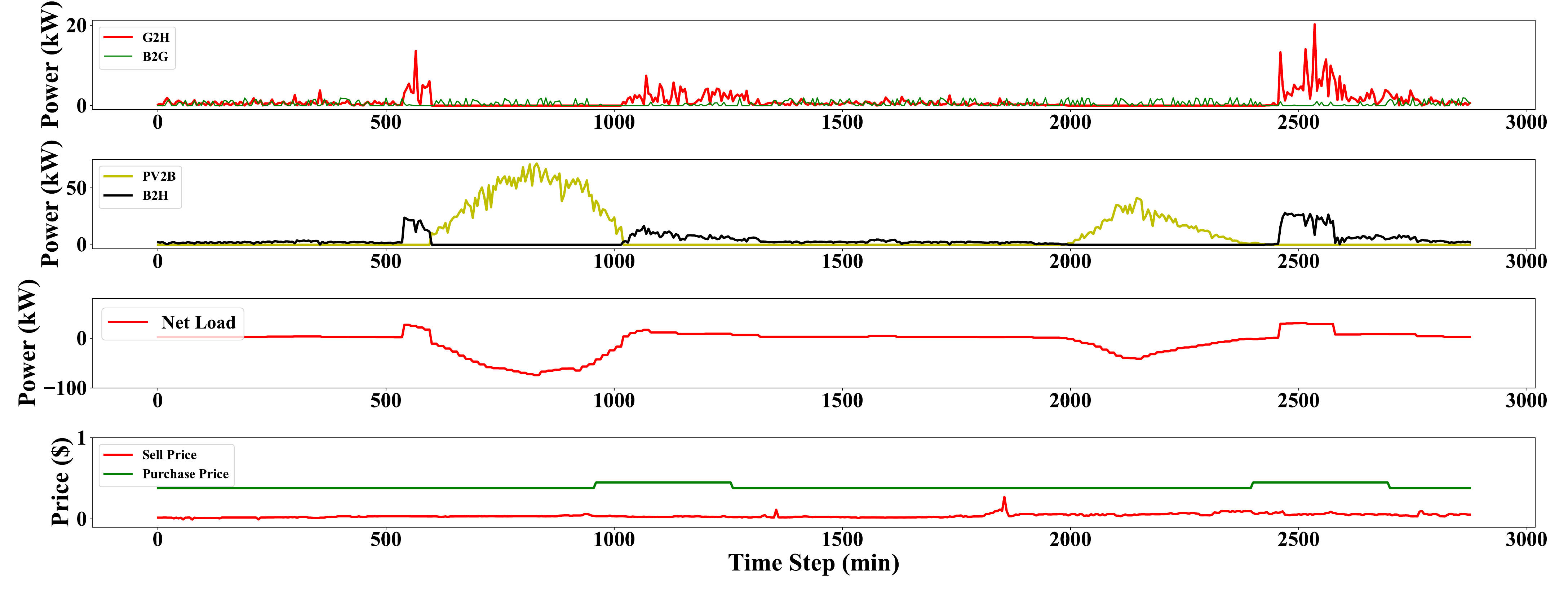}
    \caption{The visualization of the community battery's decision under uncertainties.}
    \label{comp}
\end{figure*}
Figure \ref{comp} presents the decision-making process of the community battery under uncertainties. The first and second figures show the community battery's operational decisions. The third and fourth figures depict the time-series net load and energy prices, respectively. In response to the PV-to-Battery (PV2B) strategy, the agent demonstrates the ability to learn patterns in the net load, opting to charge the battery during periods of solar energy surplus, as indicated by the net load's negative values. In contrast, the Battery-to-House (B2H) strategy reflects the agent's learning to supply energy to households by discharging the community battery when solar energy is insufficient to meet the demand, as indicated by positive values of the net load. Additionally, the community battery is shown to exploit periods of lower purchase prices, drawing power from the grid. These experimental results demonstrate the effectiveness of the SAC algorithm under various uncertainties.

The goal of the RL algorithm is to minimize the total cost and capacity constraint violations. Table \ref{pe} shows the energy cost and number of constraint violations for different algorithms. The results highlight that SAC outperforms other other algorithms in reducing energy cost and violation numbers. Though the model-based optimization can satisfy constraints, its cost is higher. The performance of PPO and DDPG is dominated by SAC.
\begin{table}[t]
\centering
\caption{Performance comparisons between model-based optimization, DDPG, PPO and SAC.}
\setlength{\tabcolsep}{0.5mm}{\begin{tabular}{||c | c | c ||} 
 \hline
 Algorithm & Energy Cost (\$) & Constraint Violation Number \\ [0.5ex] 
 \hline
 Model-based & 281.3 & 0  \\ 
 DDPG & 307.6 & 13  \\ 
 PPO & 292.3 & 6 \\
 \textbf{SAC}& 272.2 & 3  \\
 \hline
\end{tabular}}
\label{pe}
\end{table}

\section{Conclusion} \label{sec:con}
This paper demonstrated the effectiveness of DRL, in particular SAC, in scheduling a community battery system amidst uncertainties. The community battery serves effective roles in the integration of local solar PV energy, reducing peak load, and participating in energy arbitrage. We employed the SAC RL algorithm to solve the community battery scheulding problem. We also incorporated a noisy network for better exploration during RL training to enhance convergence. Our work contributes to the growing exploration of RL's potential in community battery scheduling. Our future research will consider the integration of community battery into the grid to provide ancillary services, taking into account network constraints.

\bibliographystyle{IEEEtran}
\bibliography{IEEEabrv.bib,ref.bib}

\begin{thebibliography}{10}
\providecommand{\url}[1]{#1}
\csname url@samestyle\endcsname
\providecommand{\newblock}{\relax}
\providecommand{\bibinfo}[2]{#2}
\providecommand{\BIBentrySTDinterwordspacing}{\spaceskip=0pt\relax}
\providecommand{\BIBentryALTinterwordstretchfactor}{4}
\providecommand{\BIBentryALTinterwordspacing}{\spaceskip=\fontdimen2\font plus
\BIBentryALTinterwordstretchfactor\fontdimen3\font minus \fontdimen4\font\relax}
\providecommand{\BIBforeignlanguage}[2]{{%
\expandafter\ifx\csname l@#1\endcsname\relax
\typeout{** WARNING: IEEEtran.bst: No hyphenation pattern has been}%
\typeout{** loaded for the language `#1'. Using the pattern for}%
\typeout{** the default language instead.}%
\else
\language=\csname l@#1\endcsname
\fi
#2}}
\providecommand{\BIBdecl}{\relax}
\BIBdecl

\bibitem{koirala2018community}
B.~P. Koirala, E.~van Oost, and H.~van~der Windt, ``Community energy storage: A responsible innovation towards a sustainable energy system?'' \emph{Applied energy}, vol. 231, pp. 570--585, 2018.

\bibitem{zhao2019virtual}
D.~Zhao, H.~Wang, J.~Huang, and X.~Lin, ``Virtual energy storage sharing and capacity allocation,'' \emph{IEEE transactions on smart grid}, vol.~11, no.~2, pp. 1112--1123, 2019.

\bibitem{zhao2017pricing}
------, ``Pricing-based energy storage sharing and virtual capacity allocation,'' in \emph{2017 IEEE International Conference on Communications (ICC)}.\hskip 1em plus 0.5em minus 0.4em\relax IEEE, 2017, pp. 1--6.

\bibitem{abraham2023mpc}
S.~Abraham, Y.~Mishra, and M.~E. Cholette, ``Mpc based community battery system to minimize the energy cost of a residential community,'' in \emph{2023 IEEE Power \& Energy Society General Meeting}.\hskip 1em plus 0.5em minus 0.4em\relax IEEE, 2023, pp. 1--5.

\bibitem{wolfs2012receding}
P.~Wolfs and G.~S. Reddy, ``A receding predictive horizon approach to the periodic optimization of community battery energy storage systems,'' in \emph{2012 22nd Australasian Universities Power Engineering Conference (AUPEC)}.\hskip 1em plus 0.5em minus 0.4em\relax IEEE, 2012, pp. 1--6.

\bibitem{pezeshki2014model}
H.~Pezeshki, P.~Wolfs, and G.~Ledwich, ``A model predictive approach for community battery energy storage system optimization,'' in \emph{2014 IEEE Power \& Energy Society General Meeting}.\hskip 1em plus 0.5em minus 0.4em\relax IEEE, 2014, pp. 1--5.

\bibitem{stentati2022optimization}
M.~Stentati, S.~Paoletti, and A.~Vicino, ``Optimization of energy communities in the italian incentive system,'' in \emph{2022 IEEE PES Innovative Smart Grid Technologies Conference Europe}.\hskip 1em plus 0.5em minus 0.4em\relax IEEE, 2022, pp. 1--5.

\bibitem{van2020low}
W.~van Westering and H.~Hellendoorn, ``Low voltage power grid congestion reduction using a community battery: Design principles, control and experimental validation,'' \emph{International Journal of Electrical Power \& Energy Systems}, vol. 114, p. 105349, 2020.

\bibitem{wang2017pv}
H.~Wang, N.~Good, P.~Mancarella, and K.~Lintern, ``Pv-battery community energy systems: Economic, energy independence and network deferral analysis,'' in \emph{2017 14th International Conference on the European Energy Market (EEM)}.\hskip 1em plus 0.5em minus 0.4em\relax IEEE, 2017, pp. 1--5.

\bibitem{wang2023community}
Y.~Wang, H.~Wang, M.~Wagner, and A.~Liebman, ``Community battery energy storage systems for enhancing distribution system operation: A multi-objective optimization approach,'' in \emph{2023 IEEE International Conference on Energy Technologies for Future Grids (ETFG)}.\hskip 1em plus 0.5em minus 0.4em\relax IEEE, 2023, pp. 1--6.

\bibitem{wang2018energy}
H.~Wang and B.~Zhang, ``Energy storage arbitrage in real-time markets via reinforcement learning,'' in \emph{2018 IEEE Power \& Energy Society General Meeting}.\hskip 1em plus 0.5em minus 0.4em\relax IEEE, 2018, pp. 1--5.

\bibitem{badoual2021learning}
M.~D. Badoual and S.~J. Moura, ``A learning-based optimal market bidding strategy for price-maker energy storage,'' in \emph{2021 American Control Conference (ACC)}.\hskip 1em plus 0.5em minus 0.4em\relax IEEE, 2021, pp. 526--532.

\bibitem{li2023deep}
J.~Li, C.~Wang, and H.~Wang, ``Deep reinforcement learning for wind and energy storage coordination in wholesale energy and ancillary service markets,'' \emph{Energy and AI}, p. 100280, 2023.

\bibitem{sultanuddin2023development}
S.~Sultanuddin, R.~Vibin, A.~R. Kumar, N.~R. Behera, M.~J. Pasha, and K.~Baseer, ``Development of improved reinforcement learning smart charging strategy for electric vehicle fleet,'' \emph{Journal of Energy Storage}, vol.~64, p. 106987, 2023.

\bibitem{cao2021smart}
Y.~Cao, H.~Wang, D.~Li, and G.~Zhang, ``Smart online charging algorithm for electric vehicles via customized actor--critic learning,'' \emph{IEEE Internet of Things Journal}, vol.~9, no.~1, pp. 684--694, 2021.

\bibitem{alfaverh2023optimal}
F.~Alfaverh, M.~Dena{\"\i}, and Y.~Sun, ``Optimal vehicle-to-grid control for supplementary frequency regulation using deep reinforcement learning,'' \emph{Electric Power Systems Research}, vol. 214, p. 108949, 2023.

\bibitem{kosuru2021reinforcement}
R.~Kosuru, S.~Liu, and H.~Chaoui, ``A reinforcement learning based energy management system for a pv and battery connected microgrid system,'' in \emph{2021 IEEE 30th International Symposium on Industrial Electronics (ISIE)}.\hskip 1em plus 0.5em minus 0.4em\relax IEEE, 2021, pp. 01--05.

\bibitem{rostmnezhad2023power}
Z.~Rostmnezhad and L.~Dessaint, ``Power management in smart buildings using reinforcement learning,'' in \emph{2023 IEEE Power \& Energy Society Innovative Smart Grid Technologies Conference}.\hskip 1em plus 0.5em minus 0.4em\relax IEEE, 2023, pp. 1--5.

\bibitem{liu2023learning}
R.~Liu and Y.~Chen, ``Learning a multi-agent controller for shared energy storage system,'' in \emph{2023 IEEE Power \& Energy Society General Meeting}.\hskip 1em plus 0.5em minus 0.4em\relax IEEE, 2023, pp. 1--5.

\bibitem{lix2023deep}
X.~Li, H.~Liu, C.~Li, G.~Chen, C.~Zhang, and Z.~Y. Dong, ``Deep reinforcement learning based explainable pricing policy for virtual storage rental service,'' \emph{IEEE Transactions on Smart Grid}, 2023.

\bibitem{fan2023marl}
J.~Fan, H.~Wang, and A.~Liebman, ``Marl for decentralized electric vehicle charging coordination with v2v energy exchange,'' in \emph{The 49th Annual Conference of the IEEE Industrial Electronics Society (IECON)}.\hskip 1em plus 0.5em minus 0.4em\relax IEEE, 2023, pp. 1--6.

\bibitem{wu2022optimal}
Y.~Wu, Z.~Liu, J.~Liu, H.~Xiao, R.~Liu, and L.~Zhang, ``Optimal battery capacity of grid-connected pv-battery systems considering battery degradation,'' \emph{Renewable Energy}, vol. 181, pp. 10--23, 2022.

\bibitem{haarnoja2018soft}
T.~Haarnoja, A.~Zhou, P.~Abbeel, and S.~Levine, ``Soft actor-critic: Off-policy maximum entropy deep reinforcement learning with a stochastic actor,'' in \emph{International conference on machine learning}.\hskip 1em plus 0.5em minus 0.4em\relax PMLR, 2018, pp. 1861--1870.

\bibitem{plappert2017parameter}
M.~Plappert, R.~Houthooft, P.~Dhariwal, S.~Sidor, R.~Y. Chen, X.~Chen, T.~Asfour, P.~Abbeel, and M.~Andrychowicz, ``Parameter space noise for exploration,'' \emph{arXiv preprint arXiv:1706.01905}, 2017.

\bibitem{aemo_dashboard}
AEMO, ``Nem data dashboard,'' 2023, https://aemo.com.au/energy-systems/electricity/national-electricity-market-nem/data-nem/data-dashboard-nem.

\end{thebibliography}

\end{document}